\documentclass[runningheads]{llncs}
\usepackage[T1]{fontenc}
\usepackage{graphicx}
\usepackage[dvipsnames]{xcolor}

\usepackage{graphicx}       %载入图形包
\usepackage{diagbox}
\usepackage{multirow}
\newcommand{\cF}{\mathcal{F}}
\newcommand{\cG}{\mathcal{G}}

\newcommand{\cN}{\mathcal{N}}

\makeatletter
\DeclareRobustCommand\onedot{\futurelet\@let@token\@onedot}
\def\@onedot{\ifx\@let@token.\else.\null\fi\xspace}

\makeatother

\definecolor{darkgreen}{rgb}{0,0.7,0}

\usepackage[utf8]{inputenc}
\usepackage{times}    % Integrate Times for here
\usepackage{cite}     % Automatically ordered citations
\usepackage{xspace}
\usepackage{graphicx}
\usepackage{amsmath}
\usepackage{amssymb}
\usepackage{booktabs}
\usepackage{enumitem}
\usepackage{enumerate}
\usepackage{cleveref}

\begin{document}
\title{\vspace{-0.2cm} Rethinking Attention Module Design for \\ Point Cloud Analysis \vspace{-0.2cm}}

\author{
% \small
Chengzhi Wu$^{1}$ \quad Kaige Wang$^{1}$ \quad Zeyun Zhong$^{1}$ \quad Hao Fu$^{1}$ \quad Junwei Zheng$^{1}$ \\
Jiaming Zhang$^{1}$ \quad Julius Pfrommer$^{2}$ \quad Jürgen Beyerer$^{1,2}$ \vspace{-0.1cm}
}
% \vspace{1pt}
% \and
\institute{{
\small
$^{1}$Institute for Anthropomatics and Robotics, Karlsruhe Institute of Technology, Germany \\
$^{2}$Fraunhofer Institute of Optronics, System Technologies and Image Exploitation IOSB, Germany 
% $^{1}$Karlsruhe Institute of Technology, Germany \quad
% $^{2}$Fraunhofer IOSB, Germany
% \vspace{-0.2cm}
}
% \and
{%\tt\footnotesize
\tt\fontsize{7.5pt}{6pt}\selectfont
\{chengzhi.wu, zeyun.zhong, junwei.zheng, jiaming.zhang\}@kit.edu, \vspace{-0.1cm} \\
\{kaige.wang, hao.fu\}@student.kit.edu,} \vspace{-0.1cm} \\
{%\tt\footnotesize
\tt\fontsize{7.5pt}{6pt}\selectfont
\{julius.pfrommer, juergen.beyerer\}@iosb.fraunhofer.de}
\vspace{-0.1cm}}

\authorrunning{C. Wu et al.}

\maketitle              % typeset the header of the contribution
\begin{abstract}
In recent years, there have been significant advancements in applying attention mechanisms to point cloud analysis. However, attention module variants featured in various research papers often operate under diverse settings and tasks, incorporating potential training strategies. This heterogeneity poses challenges in establishing a fair comparison among these attention module variants. In this paper, we address this issue by rethinking and exploring attention module design within a consistent base framework and settings. Both global-based and local-based attention methods are studied, with a focus on the selection basis and scales of neighbors for local-based attention. Different combinations of aggregated local features and computation methods for attention scores are evaluated, ranging from the initial addition/concatenation-based approach to the widely adopted dot product-based method and the recently proposed vector attention technique. Various position encoding methods are also investigated. Our extensive experimental analysis reveals that there is no universally optimal design across diverse point cloud tasks. Instead, drawing from best practices, we propose tailored attention modules for specific tasks, leading to superior performance on point cloud classification and segmentation benchmarks.

\end{abstract}

\section{Introduction}
\label{sec:intro}

The attention mechanism was first proposed by Bahdanau et al. \cite{bahdanau2016neural} in 2014 to learn richer information from the input. 
Later, given its remarkable performance in the natural language processing domain \cite{vaswani2017attention} and 2D computer vision of image analysis \cite{ranftl2021vision}, the research community also started to explore the application of attention modules for 3D point clouds.
In 2021, Point Cloud Transformer (PCT) \cite{guo2021pct} and $\textrm{PT}^{1}$ \cite{zhao2021point} were the first to apply the attention mechanism to the point cloud learning tasks. In the following years, several improvements have been made to the attention mechanism from different perspectives for better feature learning of point clouds \cite{shan2021ptt,han2021dual,cheng2022patchformer,qin2022geometric,pointmixswap,huang2024pointcat,hou2022hitpr,ding2022point}. 

However, methods in different papers usually run under different settings and tasks, with potential various training tricks. Many papers claim that their proposed new modules achieve better performance, but the performance improvement may possibly be obtained due to the modifications in other parts. In this case, it is hard to determine which module is actually the optimal solution for one certain task. Hence, this work aims to conduct a comprehensive study on various attention module variants and provide a more equitable comparison, then propose more effective attention-based fundamental modules for different point cloud downstream tasks. Moreover, to better investigate how one attention module variant behaves under different downstream tasks, we adopt the same network model in different tasks for a fair comparison. 

In this work, we explore the following four key aspects for attention module variants used in point cloud analysis: (1) neighbor selection operation; (2) local feature aggregation; (3) attention score computation methods; and (4) possible position encoding. Note that the former two aspects are only involved in local-based attention.

In neighbor selection operation for local-based learning, the measure of "distance" between points is mostly based on the coordinate distance or the feature difference between points. 
Apart from single-scale neighbor grouping, a multi-scale grouping strategy was also adopted in some papers. For example, Stratified Transformer \cite{lai2022stratified} selects multi-scale neighbors through a stratified strategy for key sampling, while 3DCTN \cite{lu20223dctn} implements multi-scale neighbor selection via a parallel multi-level multi-scale point transformer.
For local feature aggregation, a certain combination of the following three types of features are often used for feature aggregation: features of the centers, features of the neighbors, and the feature difference between the neighbor points and the center points \cite{wang2019dynamic}. 

Apart from the widely used dot product operation for computing attention scores, there are many other operation choices, including addition, concatenation, and subtraction. Each one also has more variants when considering global or local features.
For example, PCT \cite{guo2021pct} adopts the commonly used Q K dot production to compute attention scores by only considering the global information, while $\textrm{PT}^{1}$ \cite{zhao2021point} uses Q K subtraction as the first step in local feature aggregation. 
The additive-based method was used in the original attention paper from Bahdanau et al. \cite{bahdanau2016neural}. Another related research of Attention-based Neural Machine Translation \cite{luong2015effective} compared three attention approaches including multiplication, concatenation, and generalization.

Unlike text and images, point clouds are usually unordered, and thus the traditional position encoding should not be used to obtain better order variance of point inputs. 
In point cloud learning, position encoding simply means merging more information from the points' 3D coordinates directly to the attention modules.
For example, HiTPR \cite{hou2022hitpr} splices the difference between the coordinates of center and neighbor points with the difference of features, and then adds them to the attention map. LCPFormer \cite{huang2023lcpformer} projects the original coordinates of the points to the same dimension as the structural information through MLP, and then adds them with the aggregated features. ProxyFormer \cite{li2023proxyformer} uses a self-attention operation to stitch the coordinates of local points with features and send them to the attention layer.

In this work, we summarize our core technical contributions concerning attention module design for point cloud analysis as follows:
\begin{itemize}[itemsep=0pt,topsep=0pt,left=0pt]
\item Selecting multi-scale neighbors as the Key input can mostly improve model performance yet model size and FLOPs increase significantly. To improve model performance with the same number of neighbors, grouping neighbors with a skipping strategy to achieve a larger perceptive field is recommended.
\item In local feature aggregation, using the offset feature (the difference between the center point feature and the neighbor point feature) mostly yields a better result compared to using the neighbor feature directly.
\item For global-based attention, L2-norm subtraction-based attention is overall better than the dot product self-attention. For local-based attention, offset-based attention modules achieve relatively better performance in both the scalar and vector attention cases.
\item Applying implicit position encoding is better than explicitly concatenating point coordinates to the attention input. Most implicit position encoding methods achieve similar favorable outcomes under various attention methods, and compressing its feature dimension during the encoding leads to less improved performance.
\item We reveal that there is no such attention module that always achieves the best performance under different downstream tasks. However, some insights for choosing an optimal one are given through our exploration.
\end{itemize}

\section{Related Work}
\label{sec:relatedWork}
%This section will center around getting valid point cloud features as inputs to the attention layer, the attention mechanism itself, and the positional encoding methods combined with the attention mechanism.\\

\textbf{Point Cloud Local Feature Aggregation.} 
The pioneering work by Qi et al. introduced PointNet \cite{qi2017pointnet}, which introduces global attention mechanism in point cloud processing. Building upon this, PointNet++ \cite{qi2017pointnet++} was introduced, incorporating local feature aggregation through a hierarchical neural network that captures fine-grained patterns and global context. Wang et al. proposed DGCNN \cite{wang2019dynamic}, leveraging edge convolution to aggregate local features based on the k-nearest neighbors in the feature space. HiTPR \cite{hou2022hitpr}  leverages local feature aggregation to enhance the model's ability to recognize and match places in 3D environments. KPConv \cite{thomas2019kpconv} introduces a deformable convolution operation for more flexible processing of irregularly distributed point cloud data. PointCNN \cite{li2018pointcnn} A novel X-convolution operation is proposed to better adapt to the disorder point clouds. PPTNet \cite{hui2021pyramid} utilizes a transformer-based architecture to perform a variety of point cloud processing tasks, which can capture the ability of fine-grained details and broader contextual information \cite{yu2022point,lai2022stratified}. MLMSPT \cite{zhong2021point}, 3DGTN \cite{lu20223dctn} uses a multi-scale neighbor point selection method to establish receptive fields of different scales and densities for extracting local features. 

\textbf{Attention-based Deep Learning on Point Clouds.}
By introducing the attention mechanism\cite{paigwar2019attentional,feng2020point}, models have made great progress in point cloud detection and segmentation tasks. By using the attention mechanism, the model can selectively focus on important points or regions, effectively processing large-scale and irregularly sampled point clouds \cite{hu2020randla,thomas2019kpconv,wang2019associatively,liu2019point2sequence,wu2023sim2real,wu20246d}.
PCT \cite{guo2021pct} applies the Transformer model directly to point cloud classification and segmentation tasks. Cross attention has been used to aggregate features from different regions of a point cloud, leading to more accurate and fine-grained predictions \cite{zhao2021point,engel2021point,xu2021you,hou2022hitpr,xia2023casspr,trappolini2021shape,cui20213d,park2023self,wu2023attention,wu2024cross}. PTTR \cite{zhou2022pttr} divides the input point cloud into multiple groups of points to extract their features respectively and then match and predict these features. PT V2 \cite{wu2022point} incorporates more efficient attention mechanisms and network architectures to boost performance while reducing computational demands.

\textbf{Position Encoding in Point Cloud Analysis.} Regarding position encoding,  Point Transformer \cite{zhao2021point} introduces position encoding into point cloud analysis, greatly improving model performance, and PPTNet \cite{hui2021pyramid}further extends this work. Liu et al. \cite{zhang2019rotation} proposed a rotation-invariant position encoding method to ensure that the model performs stably on input point clouds in different directions. RandLA-Net \cite{hu2020randla} explores the efficiency of position encoding in large-scale point cloud applications. Kan Wu et al. comparatively studied the effects of absolute position encoding \cite{vaswani2017attention} and relative position \cite{huang2020improve,ramachandran2019stand,shaw2018self,wang2020axial}under the attention mechanism, And a contextual-based position encoding method is proposed \cite{wu2021rethinking}, this method and its variants have been widely used in point clouds \cite{wu2022point, qin2022geometric, zhou2022seedformer, lai2022stratified}.

\section{Attention Module Variants}
\label{sec:methodology}

\begin{figure}[t]
\centerline{\includegraphics[width=0.8\linewidth]{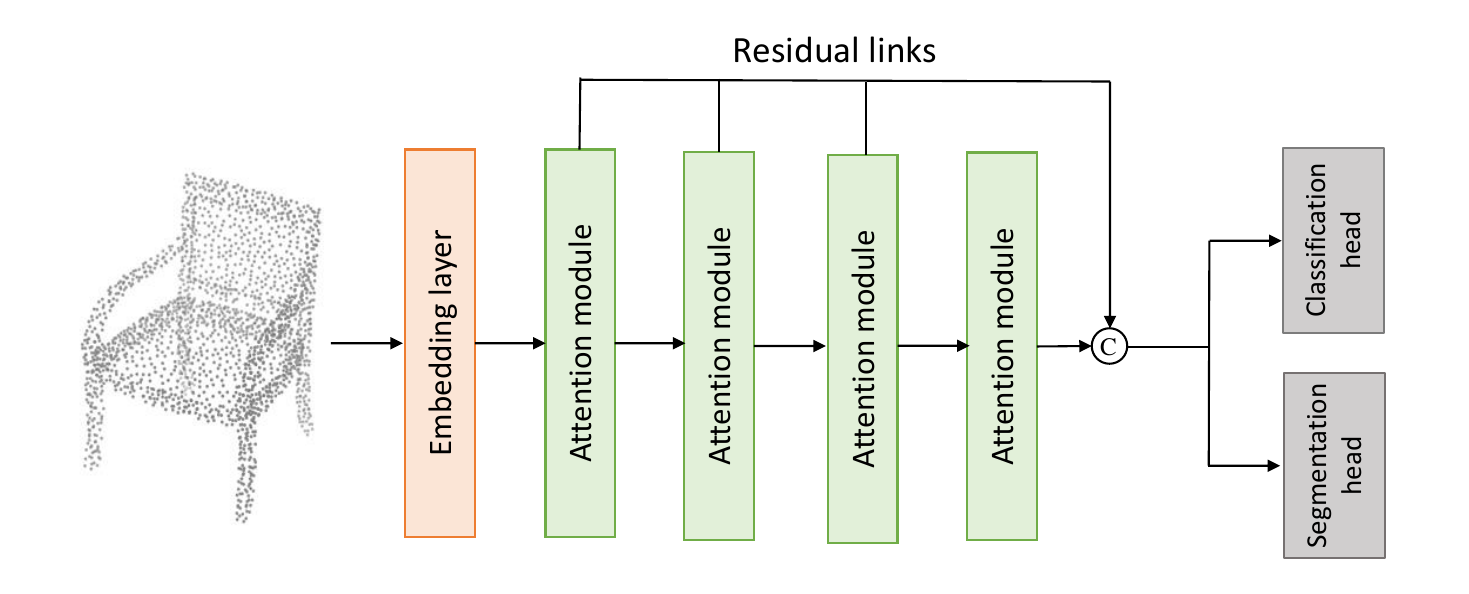}}
\caption{Basic Framework. It consists of an embedding layer, four sequential attention modules with residual links, and a task-oriented MLP head.}
\label{fig:basic_framework}
\end{figure}

Methods in different papers run under different frameworks and settings, with potential various training tricks.
In order to conduct a comprehensive study on various attention module variants and provide a more equitable comparison, we use an identical basic framework in all the experiments for one certain task, with the same setting and no special training tricks. The framework is given in \Cref{fig:basic_framework}. It consists of an embedding layer, four sequential attention modules with residual links, and a task-oriented MLP head.

%%%%%%%%%%%% figure: basic framework %%%%%%%%%%

The module differences of various attention module variants can be summarized in the following four aspects: neighbor selection, local feature aggregation, position encoding, and Q K V feature fusion method (i.e., the actual attention method). Note that the former two aspects are only involved in local-based attention.

\subsection{Neighbor Selection}
\label{sec:neighbor selection}
\begin{figure}[t]
\centering
\includegraphics[width=\linewidth,trim=0 0 0 0, clip] {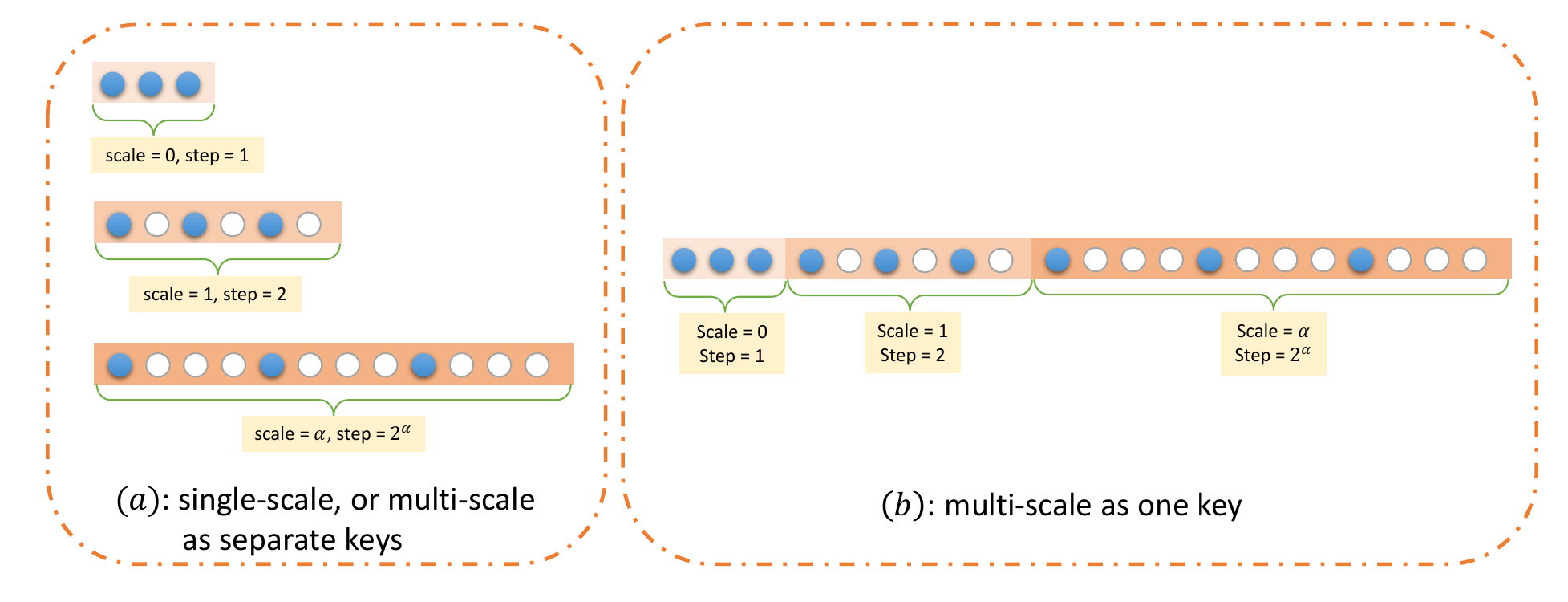}
\caption{ (a) Single-scale, or multi-scale as separate keys, and (b) multi-scale as one key. A sparser point selection method is used in larger perceptual fields, with the number of points selected in each scale being consistent.}
\label{fig:ms_one_or_sep}    
\end{figure}

When considering local information, the first decision to make is on what basis the neighbors are selected, i.e., on original point 3D coordinates, or the feature similarity in the high-dimensional feature space. After the basis is determined, $k$ neighbors are selected with the K-Nearest Neighbors (KNN) method in the vanilla case. However, it is possible to select the same number of neighbor points with a larger perceptual field by regular skipping. As illustrated in \Cref{fig:ms_one_or_sep}(a), for scale $\alpha$, $k*2^{\alpha}$ nearest neighbors are first obtained, then $k$ points are selected with a step of $2^{\alpha}$.

Moreover, it is possible to consider the case of multi-scale, i.e., use the neighbor groups of different perceptual field sizes as multiple keys for the attention operation. 
For multi-scale as separate keys, the number of keys in the attention layer is equal to the number of scales. Each scale selects points from the same starting point using different degrees of sparsity. For multi-scale as one key, there is no overlapping between different scales.
As illustrated in \Cref{fig:ms_one_or_sep}(b), to avoid repetitive point selection, multi-scale as separate keys method obtains $k*(2^{\alpha+1}-1)$ nearest neighbors first, then select $k$ points with different steps in different segments.

\subsection{Local Feature Aggregation}
\label{sec:localFeaAggre}
In local-based attention, the features that can be used for local feature aggregation include (i) the feature of the center point; (ii) the feature of selected neighbor points; and (iii) the offset feature between neighbor points and the center point. 
Different combinations of these three methods are tested for different attention methods. K should include at least one of the neighbor feature and the offset feature.

%%%%%%%%%%%% figure: multi-scale %%%%%%%%%%

It is worth noting that the attention score computation method influences the choice of features used for local feature aggregation. For example, when using offset/subtraction-based attention methods to compute attention scores, the operation already contains the offset feature information implicitly (Q contains the center feature, while K contains the neighbor feature). Hence the local feature aggregation part should exclude the feature differences in this case. Another example is that when using addition-based attention methods to compute attention scores, the local feature aggregation part should exclude the center feature. A corresponding table is given in \Cref{table:local_aggre}. Detailed introductions of various attention operation methods are given in \Cref{sec:attention_method}.

\begin{table}[t]
\centering
\caption{Possible combinations of features for local feature aggregation in different attention operation methods.}
\resizebox{0.7\linewidth}{!}{
\begin{tabular}{ccccccc}
\toprule
 \diagbox{Att. method}{Local fea.} &
  neighbor &
  offset &
  \begin{tabular}[c]{@{}c@{}}center, \\ neighbor\end{tabular} &
  \begin{tabular}[c]{@{}c@{}}center, \\ offset\end{tabular} &
  \begin{tabular}[c]{@{}c@{}}neighbor, \\ offset\end{tabular} &
  \begin{tabular}[c]{@{}c@{}}center, \\ neighbor, \\offset\end{tabular} \\ \midrule
Dot product        & $\checkmark$ & $\checkmark$ & $\checkmark$ & $\checkmark$ & $\checkmark$ & $\checkmark$ \\ 
Offset/Subtraction & $\checkmark$ & - & $\checkmark$ & - & - & - \\ 
Addition           & $\checkmark$ & $\checkmark$ & - & - & $\checkmark$ & - \\ 
Concat             & $\checkmark$ & $\checkmark$ & - & - & $\checkmark$ & - \\ \bottomrule
\end{tabular}}
\label{table:local_aggre}
\end{table}

\subsection{Attention Method}
\label{sec:attention_method}
Based on whether local information is considered, the attention methods used for point cloud analysis can be mainly divided into two categories: global-based attention, and local-based attention.

\subsubsection{Global-based Attention} 
\begin{figure}[t]
    \centering
    \includegraphics[width=0.9\linewidth,trim=10 0 10 0, clip]
    {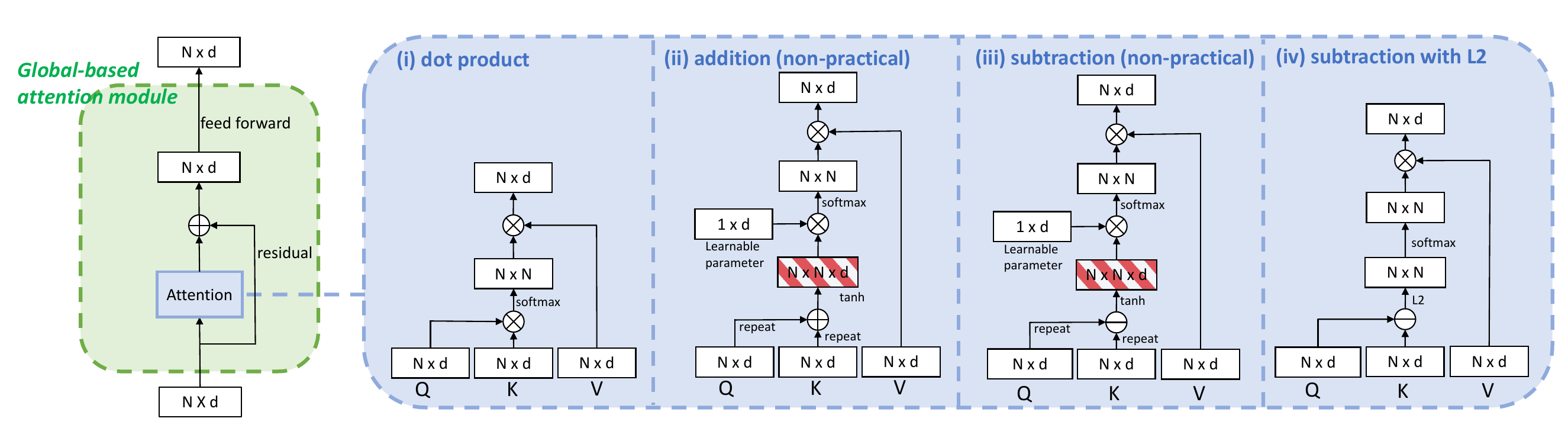}
    \caption{Global-based attention module.}
    \label{fig:global_based_attention_module}
\end{figure}

Global-based attention module is designed as shown in \Cref{fig:global_based_attention_module}. It can be described as $\cF_g$:
\begin{equation}
    \cF_g = \text{FFN}(\phi^{g}(p) + p)
    \label{eq:eq_global}
\end{equation}
where $\phi^{g}$ denotes the global self-attention function, and $p$ represents the point cloud features and is the input of the entire module. A residual link is used to convey the input to the post-attention tensor.
Finally, the output is obtained through a feedforward neural network (FFN).

\begin{figure*}[t]
    \centering
    \includegraphics[width=\linewidth,trim=0 0 0 0, clip] {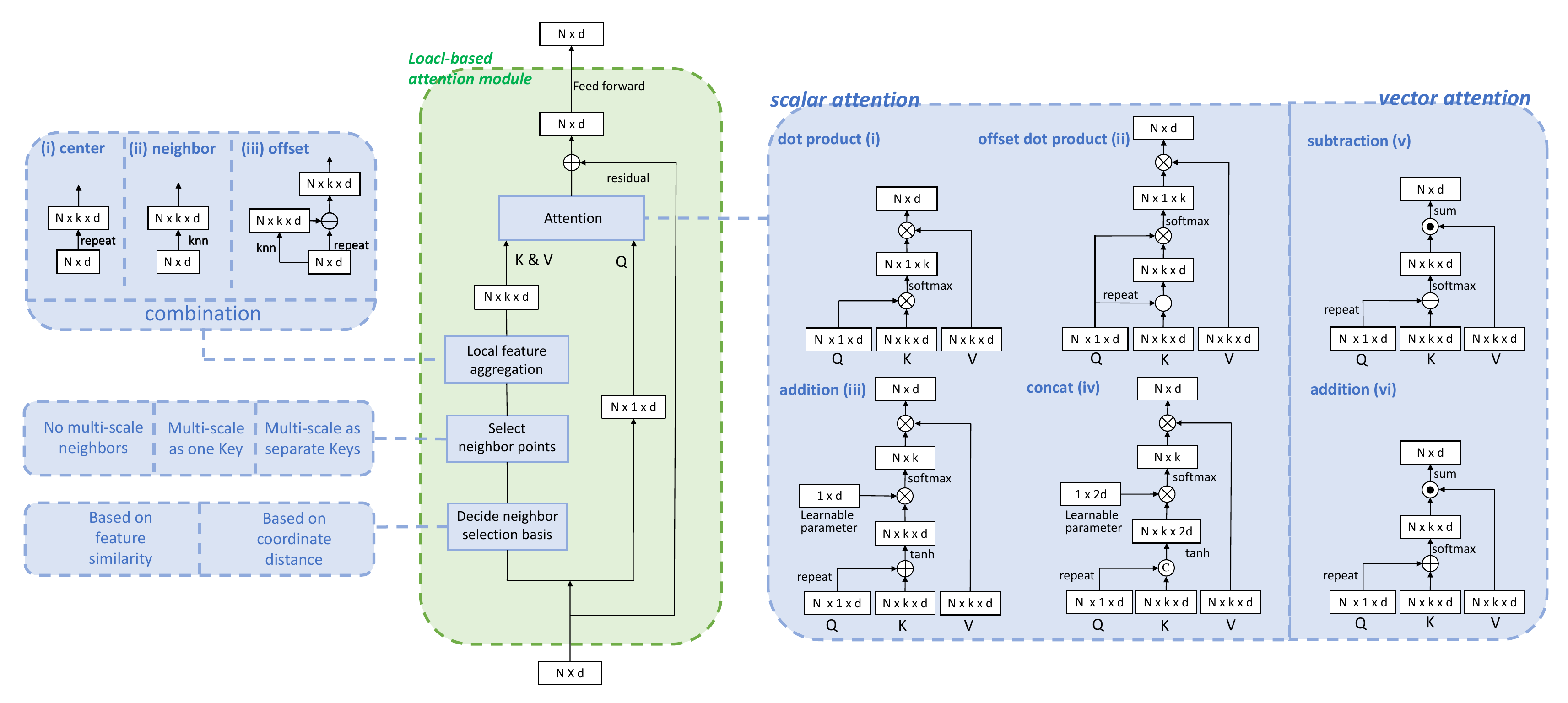}
    \caption{Local-based attention module.}
    \label{fig:local_based_attention_module}
\end{figure*}

\textbf{Dot Product.}
For global-based attention modules, $\phi^{g}$ in \Cref{eq:eq_global} is regarded as the main research object. 
Q K dot product as shown in \Cref{fig:global_based_attention_module}(i) is widely used as a calculation method for attention scores\cite{shan2021ptt,han2021dual,cheng2022patchformer,qin2022geometric,pointmixswap,huang2024pointcat}, and it can be described by the following function:
\begin{equation}
   \phi^{g}_{dot} = \text{Softmax}\left(\frac{Q \cdot K}{\sqrt{d}}\right) \cdot V
\label{eq:golbal_dot_product}    
\end{equation}
\hspace{1em}
\textbf{Direct Addition and Subtraction.}
The computational methods of addition \cite{bahdanau2016neural} and subtraction %\cite{chen2022psformer, hou2022hitpr} 
can also be used to compute attention scores. However as shown in \Cref{fig:global_based_attention_module} (ii) and (iii), the problem of too large size tensor ($N \times N \times d$) arises in the computational procedure, which makes the methods actually inapplicable. 

\textbf{Subtraction with L2.}
On the other hand, for global attention, L2 norm-based subtraction is applicable by employing mathematical equivalence calculation tricks.
PS-Former \cite{ding2022point} use of subtraction with L2 distance to realize the calculation of attention scores, reduces the transmission dimension of the subtraction operation. The subtraction combined with L2 shown in \Cref{fig:global_based_attention_module}(iv) is described as:
\begin{equation}
    \phi^{g}_{L2} = \text{softmax}(\frac{-||Q-K||_2^2}{\sqrt{d}})\cdot V
\label{eq:eq3}
\end{equation}

\subsubsection{Local-based Attention}
Local-based attention module is designed as follows shown in \Cref{fig:local_based_attention_module}, It can be described as $\cF_l$:
\begin{align}
    \cF_l = \text{FFN}(\phi^{l} (p, \cG(p, \cN(p))+p)
    \label{eq:eq_local}
\end{align}
where  $\cN$ denotes the neighbor selection method, $\cG$ denotes the local feature aggregation method, and $\phi^{l}$ represents the local feature-based cross-attention.

For local-based attention module, neighbor selection function $\cN$ in \Cref{eq:eq_local} will be based on the difference of features between points $\cN^{f}$ or the difference of coordinates $\cN^{c}$. Neighbor points are selected based on KNN. For each point, When it is used as the center point for neighbor selection, neighbor points of multi-scale could possibly be selected. The multi-scale can be divided into two structures, multi-scale as one key $\cN_{one}$ shown in \Cref{fig:ms_one_or_sep}(a) and multi-scale as separate keys $\cN_{sep}$  shown in \Cref{fig:ms_one_or_sep}(b).

For local feature aggregation functions $\cG$, as shown in \Cref{fig:local_based_attention_module} local feature aggregation block. the features that can be used for local feature aggregation include (i) the feature of center points; (ii) the feature of k neighbor points; and (iii) the feature difference between neighbor points and center points. we will test different combinations of these three methods. It is worth noting that there is an influence between the choice of aggregation methods and the attention score computation method.

\textbf{Dot Product.} 
For cross attention function $\phi^{l}$, similar to global attention, dot product shown in \Cref{fig:local_based_attention_module}(i) are also widely used in local-based cross attention score computation:
\begin{equation}
   \phi^{l}_{dot'} = \text{softmax}\left(\frac{Q \cdot K}{\sqrt{d}}\right) \cdot V
\label{eq:local dot Product(a)}    
\end{equation}

\textbf{Offset dot Product.} 
For the dot product method, if the input is the difference between neighbor points and center points(offset feature), the attention module performs the “offset before MLP” operation. In the meta-base model \cite{lin2023meta}, Lin et al. completed a comparative experiment on whether to perform MLP before Group \cite{zhao2021point,lai2022stratified,li2018pointcnn,liu2020closer} or Group before MLP \cite{li2019deepgcns, qian2022pointnext, qi2017pointnet, wang2019dynamic}. Inspired by this experiment, as a comparison with dot product, dot product with offset shown in \Cref{fig:local_based_attention_module}(ii) is based on the dot product setting of MLP before offset:

\begin{equation}
   \phi^{l}_{dot''} = \text{softmax}\left(\frac{Q \cdot (Q_r-K)}{\sqrt{d}}\right) \cdot V
\label{eq:local dot Product(b)}    
\end{equation}

\textbf{Addition.} 
When attention mechanism was proposed, the computational method of addition was used. we follow this method as one of the attention variants. \cite{bahdanau2016neural}. addition method as shown in \Cref{fig:local_based_attention_module}(iii) \cite{bahdanau2016neural} is given:\\
\begin{equation}
   \phi^{l}_{add} = \text{softmax}(\boldsymbol{\omega}^\top \cdot tanh(Q_r+K))\cdot V
\label{eq:local_addition}    
\end{equation}
where $Q_r$ means that Q repeats K times in the dimension of the number of neighbor points, $\omega^\top$ is a learnable parameter, which enhances the expressive ability of the model to a certain extent while making parameter transmission smoother and reduces information loss.\\

\textbf{Concatenation.} Similarly, using the concatenation operation on Q K and implementing the concatenation method as shown in \Cref{fig:local_based_attention_module}(iv) with the help of the learnable parameter $w$, it can be expressed as:
\begin{equation}
   \phi^{l}_{cat} = \text{softmax}(\boldsymbol{\omega}^\top \cdot tanh(concat(Q_r, K)))\cdot V
\label{eq:local_concat}    
\end{equation}

\textbf{Vector attention with subtraction.} 
Inspired by the use of vector attention mechanisms in image processing \cite{zhao2020exploring}, PT2\cite{wu2022point} uses advanced vector attention mechanisms in point clouds. Considering the rationality of feature dimension transformation, we designed vector attention functions based on subtraction shown in \Cref{fig:local_based_attention_module}(v) and addition shown in \Cref{fig:local_based_attention_module}(vi), which can be described as:\\
\begin{equation}
   \phi^{l}_{v-sub} = \text{softmax}(Q_r-K)\odot V
\label{eq:vector_substraction}    
\end{equation}
where $Q_r$ means that Q repeats $k$ times in the dimension of the number of neighbor points, $\odot$ is the Hadamard product.\\

\textbf{Vector attention with addition.} 
Since vector attention is less explored in the point cloud deep learning, we additionally test the following addition-based vector attention method:
\begin{equation}
   \phi^{l}_{v-add} = \text{softmax}(Q_r+K)\odot V
\label{eq:vector_add}    
\end{equation}

In the following experiments, the above attention score computation methods are tested with appropriate aggregated features, as listed in \Cref{table:attMethod_localFea}.

\begin{figure}[t]
    \centering
    \includegraphics[width=0.8\linewidth,trim=0 0 0 0, clip] {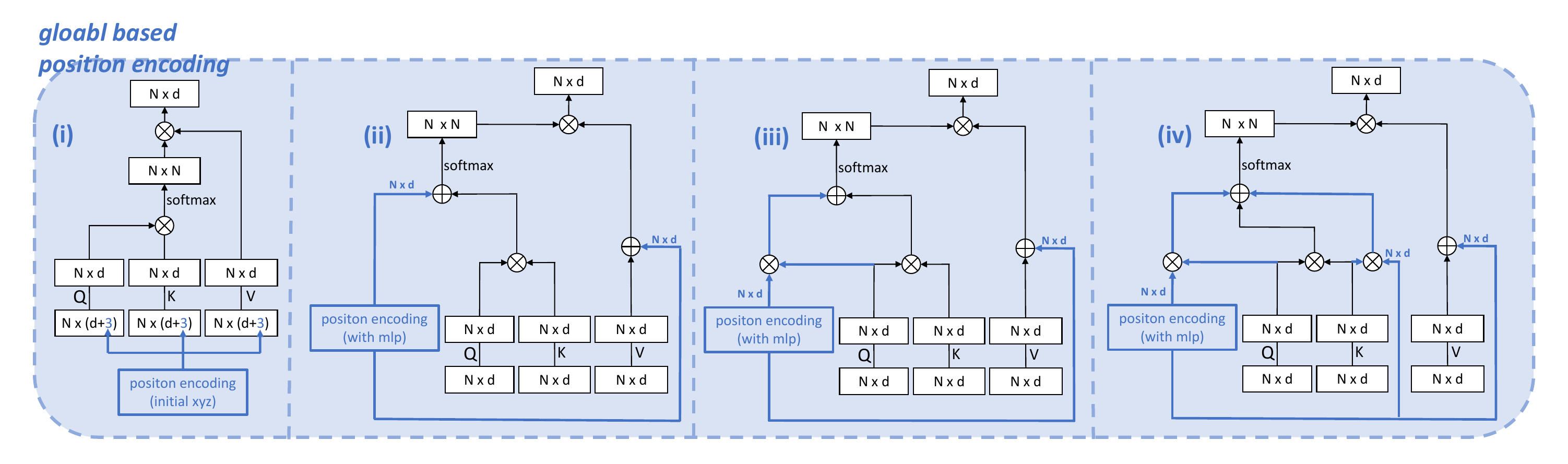}
    \caption{Position encoding in global-based attention.}
    \label{fig:global_based_position_encoding}
\end{figure}

\begin{figure}[t]
    \centering
    \includegraphics[width=0.8\linewidth,trim=0 0 0 0, clip] {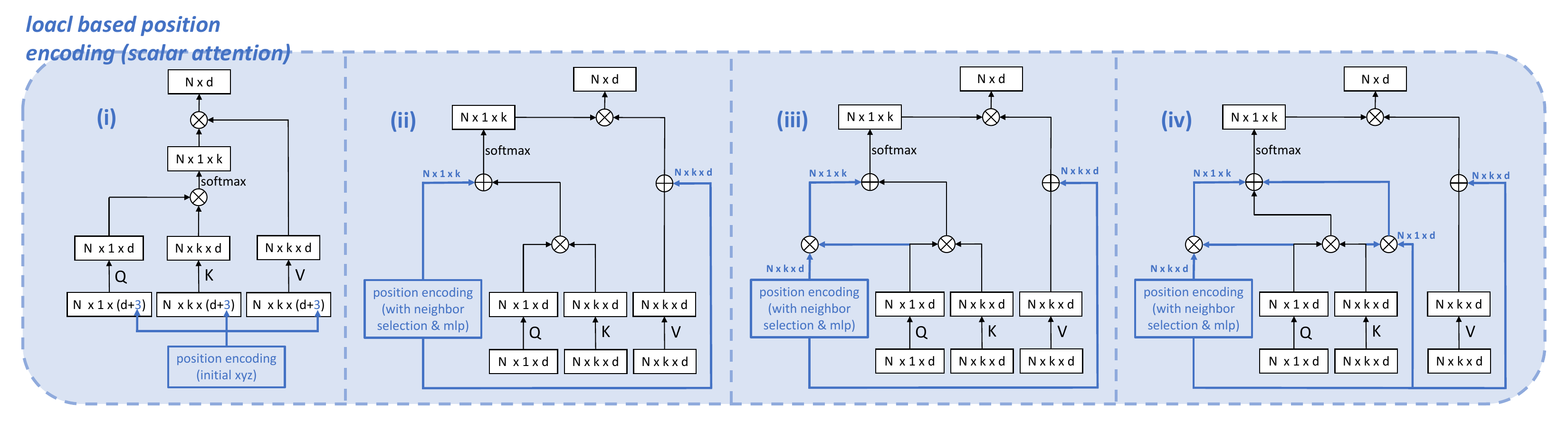}
    \caption{Position encoding in local-based scalar attention.}
\label{fig:local_based_position_encoding_scalar_attention)}
\end{figure}

\begin{figure}[t]
    \centering
    \includegraphics[width=0.7\linewidth,trim=0 0 0 0, clip] {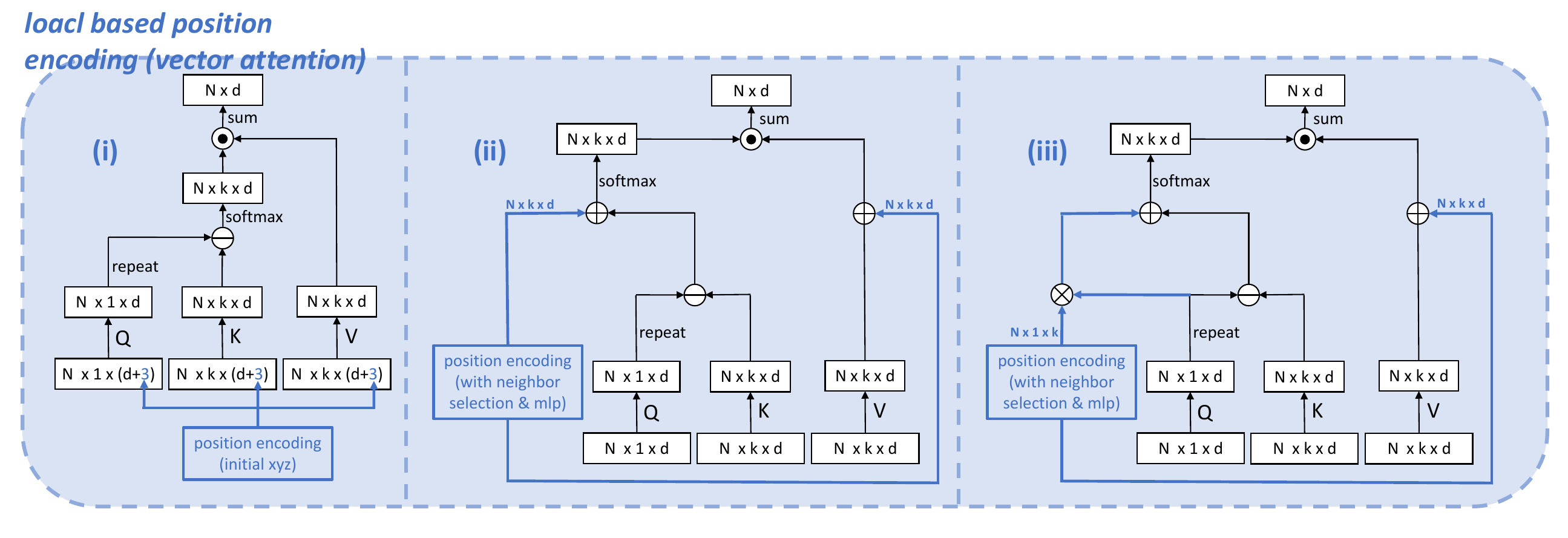}
    \caption{Position encoding in local-based vector attention.}
\label{fig:local_based_position_encoding_vector_attention)}
\end{figure}

\subsection{Position Encoding}
We explore four widely-used position encoding methods, as illustrated in \Cref{fig:global_based_position_encoding}. The first one explicitly concatenates the point coordinates with the learned latent representation from the last layer. The other three methods learn separate MLP projections and add the information to certain joints of attention modules.

\textbf{method (i).}
Based on the original self-attention considers the absolute position \cite{vaswani2017attention}, method (i) $\delta_1$ directly concatenates the spatial coordinates (x, y, z) to the Q, K, V input. By doing so, it ensures that the spatial information is explicitly merged with the attention input, thereby enhancing the network's ability to leverage absolute position information.
% \begin{equation}
%    \delta_1 = x_i + p_{xyz}
% \label{eq:position_encoding1}    
% \end{equation}

\textbf{method (ii).}
Method (ii) introduces an implicit encoding with an MLP that encodes positional information. This method relies on the network's ability to infer spatial relationships implicitly:
\begin{equation}
  \delta_2 = (Q\cdot K + \text{MLP}(p_{xyz})) \cdot (V + \text{MLP}(p_{xyz}))
\label{eq:position_encoding2}    
\end{equation}

\textbf{method (iii).}
Method (iii) enhances the implicit positional encoding by integrating it with the Q matrix. The encoding is designed to include contextual information derived from the data points' relationships in the Q space. This method uses an MLP to process the initial positional information:
\begin{equation}
   \delta_3 = (Q\cdot K + Q\cdot \text{MLP}(p_{xyz})) \cdot (V + \text{MLP}(p_{xyz}))
\label{eq:position_encoding3}    
\end{equation}

\textbf{method (iv).}
Method (iv) goes a step further by incorporating contextually enriched positional information into both the Q, K matrices. The MLP here processes the positional information and integrates it with Q and K, facilitating a more detailed comparison between points during the attention calculation:
\begin{align}
\delta_4 &= (Q \cdot K + Q \cdot \text{MLP}(p_{xyz}) + K \cdot \text{MLP}(p_{xyz})) \nonumber \\
     &\quad \cdot (V + \text{MLP}(p_{xyz}))
\end{align}

The position encoding operation under the local attention mechanism is shown in \Cref{fig:local_based_position_encoding_scalar_attention)} and \Cref{fig:local_based_position_encoding_vector_attention)}. There are differences in its dimensional transformation and global attention contrast, which will affect the performance of position encoding. This will be discussed in detail in \Cref{sec:Position_Encoding}.

\section{Explore Best Practices for Different Tasks}
\label{sec:explore_best_practices}

\subsection{Experiment setting}
\label{sec:Experiment_setting}
In this paper, two widely-used tasks are used as benchmarks for the experiments of attention module variants: point cloud classification on ModelNet40 \cite{wu20153d}, and point cloud segmentation on ShapeNetPart \cite{yi2016scalable}.

\textbf{Datasets} ModelNet40 dataset contains 12,311 pre-aligned shapes from 40 categories, which are split into 9,843 (80\%) for training and 2,468 (20\%) for testing.
ShapeNetPart dataset contains 16,881 pre-aligned shapes from 16 categories, annotated with 50 segmentation parts in total. Most object categories are labeled with two to five segmentation parts. There are 12,137 (70\%) shapes for training, 1,870 (10\%) shapes for validation, and 2,874 (20\%) shapes for testing.

\textbf{Training Details} For both classification and segmentation experiments, our setup involved training 200 epochs with a batch size of 16, splitting over 2x GTX 2080Ti for standard tasks, or 2x RTX 3090 for some multi-scale configuration experiments which need a larger memory. AdamW is used as the optimizer. The learning rate starts from $1 \times 10^{-4}$ and decays to $1 \times 10^{-8}$ with a cosine annealing schedule. A warp-up strategy is used for more stable performance.
The weight decay for model parameters is set as 1 for classification experiments and $1 \times 10^{-4}$ for segmentation experiments. We augmented the inputs by randomly applying the following four methods: jittering, rotation, translation, and anisotropic scaling.

\subsection{Neighbor Selection}
As shown in \Cref{table:neighborBasis_scale}, different multi-scale selection strategies are applied when selecting neighbor points. In this part of the experiment, we set the attention score calculation method to dot product, and the aggregation feature is the offset feature. %This method is only valid for local-based attention modules.

First, an appropriate number of neighbor points $k$ needs to be determined under different tasks. We tested the performance of the model with different numbers of neighbor points as shown in \Cref{table:k}. In the classification task, $k=32$ obtained the best performance. In the segmentation task, a larger $k$ achieves better performance. Taking the computational complexity and performance into account, we use $k=32$ for most of the following experiments.
%%%%%%%%%%%%%% table of k %%%%%%%%%%%%%%%
\begin{table}[t]
\centering
\caption{Classification and segmentation performance of using different $k$ with all other settings consistent. The applied attention method is local dot product. The offset feature is used for local feature aggregation. Only single scale $\alpha = 0$ is used.}
\resizebox{0.85\linewidth}{!}{
\begin{tabular}{cccccccc}
\toprule
\multicolumn{2}{c}{k} & 4 & 8 & 16 & 32 & 64 & 128 \\ \midrule
Cls. & OA (\%) & 92.49±0.11 & 92.74±0.09 & 92.89±0.11 & \textbf{93.10±0.06} & 92.97±0.08 & 92.84±0.06 \\ \midrule
 & Cat. mIoU (\%) & 84.04±0.08 & 84.90±0.10 & 85.53±0.08 & 85.59±0.06 & 85.90±0.07 & \textbf{86.29±0.06} \\
\multirow{-2}{*}{Seg.} & Ins. mIoU (\%) & 79.58±0.10 & 80.84±0.06 & 82.03±0.07 & 82.60±0.06 & 83.10±0.05 & \textbf{84.05±0.07} \\ \bottomrule
\end{tabular}}
\label{table:k}
\end{table}

\begin{table}[t]
\centering
\caption{Classification and segmentation performance of different variants under different neighbor selection basis and scales. The parameters and FLOPs of the attention module are also reported.}
\resizebox{0.78\linewidth}{!}{
\begin{tabular}{cccccccc}
\toprule
\toprule
\multicolumn{3}{c}{Variants} & \multirow{2}{*}{\begin{tabular}[c]{@{}c@{}}\rule{0pt}{4ex} Cls.\\ OA (\%)\end{tabular}} & \multicolumn{2}{c}{Seg.} & \multirow{2}{*}{\begin{tabular}[c]{@{}c@{}}\rule{0pt}{4ex} Params\\ (k)\end{tabular}} & \multirow{2}{*}{\begin{tabular}[c]{@{}c@{}}\rule{0pt}{4ex} FLOPs\\ (G)\end{tabular}} \\ \cmidrule(r){1-3} \cmidrule(lr){5-6}
\begin{tabular}[c]{@{}c@{}}Neighbor\\ Basis\end{tabular} & Scale & \begin{tabular}[c]{@{}c@{}}One/sep.\\ Keys\end{tabular} &  & \begin{tabular}[c]{@{}c@{}}Cat. mIoU\\ (\%)\end{tabular} & \begin{tabular}[c]{@{}c@{}}Ins. mIoU\\ (\%)\end{tabular} &  &  \\ \midrule
\multirow{7}{*}{\begin{tabular}[c]{@{}c@{}}Feature\\ Similarity\end{tabular}} & 0 & one & 93.10±0.09 & 82.60±0.06 & 85.59±0.07 & 180.22 & 2.45 \\
 & 1 & one & 93.22±0.07 & 82.87±0.08 & 85.74±0.04 & 180.22 & 2.45 \\
 & 2 & one & 93.30±0.07 & 83.02±0.07 & 85.92±0.11 & 180.22 & 2.45 \\
 & 0, 1 & one & 93.27±0.06 & 82.97±0.05 & 85.95±0.02 & 180.22 & 4.60 \\
 & 0, 1 & sep & 93.25±0.05 & 82.90±0.05 & 85.85±0.06 & 229.38 & 4.63 \\
 & 0, 1, 2 & one & \textbf{93.34±0.07} & 82.96±0.09 & \textbf{85.97±0.09} & 180.22 & 6.75 \\
 & 0, 1, 2 & sep & \textbf{93.34±0.04} & \textbf{83.15±0.06} & 85.76±0.10 & 278.53 & 6.81 \\ \midrule
\multirow{7}{*}{\begin{tabular}[c]{@{}c@{}}Initial 3D\\ Coordinates\end{tabular}} & 0 & one & 92.32±0.09 & 82.26±0.10 & 84.22±0.03 & 180.22 & 2.45 \\
 & 1 & one & 92.54±0.06 & 82.41±0.09 & 85.05±0.05 & 180.22 & 2.45 \\
 & 2 & one & 92,70±0.08 & 82.51±0.05 & \textbf{85.29±0.07} & 180.22 & 2.45 \\
 & 0, 1 & one & 92.75±0.03 & 82.55±0.06 & 85.21±0.06 & 180.22 & 4.60 \\
 & 0, 1 & sep & 92.74±0.06 & 82.54±0.06 & 85.13±0.07 & 229.38 & 4.63 \\
 & 0, 1, 2 & one & \textbf{92.93±0.07} & 82.78±0.07 & 85.25±0.06 & 180.22 & 6.75 \\
 & 0, 1, 2 & sep & 92.90±0.05 & \textbf{82.96±0.03} & 85.19±0.04 & 278.53 & 6.81  \\ \bottomrule \bottomrule
\end{tabular}}
\label{table:neighborBasis_scale}
\end{table}

As shown in \Cref{table:neighborBasis_scale}, multi-scale as one key with scales $\alpha= \{0, 1, 2\}$ achieves the best results in downstream tasks. a larger-scale point selection method can provide richer contextual information, which can help the model better generalize to different scenarios and conditions and improve model performance. On the other hand, the strategy of using separate keys does not exhibit any discernible impact, yet it requires much more model parameters. Hence we can conclude that using separate attention modules for separate scales (e.g. 3DCTN \cite{lu20223dctn}) is not necessary, using multi-scale as one key is an overall better solution (e.g. Stratified Transformer \cite{lai2022stratified}).
Moreover, if certain limitations of computational resource pose, single scale yet with a larger perceptive field (larger scale $\alpha$) for neighbor searching is also a practical solution to reduce the FLOPs of the model while still achieving decent performance.

\subsection{Local Feature Aggregation and Attention Method}
\begin{table}[t]
\centering
\caption{Classification and segmentation performance of different variants under different attention score computation methods and corresponding used features for local feature aggregation.}
\resizebox{0.9\linewidth}{!}{
\begin{tabular}{ccccccccc}
\toprule
\toprule
\multicolumn{3}{c}{Variants} & \multirow{2}{*}{\begin{tabular}[c]{@{}c@{}}\rule{0pt}{4ex} Cls.\\ OA (\%)\end{tabular}} & \multicolumn{2}{c}{Seg.} & \multirow{2}{*}{\begin{tabular}[c]{@{}c@{}}\rule{0pt}{4ex} Params\\ (k)\end{tabular}} & \multirow{2}{*}{\begin{tabular}[c]{@{}c@{}}\rule{0pt}{4ex} FLOPs\\ (G)\end{tabular}} \\ \cmidrule(r){1-3} \cmidrule(lr){5-6}
\begin{tabular}[c]{@{}c@{}}Global/\\ Local\end{tabular} & Attention & Agg. &  & \begin{tabular}[c]{@{}c@{}}Cat. mIoU\\ (\%)\end{tabular} & \begin{tabular}[c]{@{}c@{}}Ins. mIoU\\ (\%)\end{tabular} &  &  \\ \midrule
\multirow{2}{*}{Global} & Dot Product & - & 93.02±0.05 & 83.07±0.08 & 85.45±0.07 & 180.22 & 0.37015 \\
 & Subtraction & - & \textbf{93.43±0.07} & \textbf{83.15±0.06} & \textbf{85.47±0.06} & 180.22 & 0.37015 \\ \midrule
\multirow{19}{*}{Local} & \multirow{6}{*}{\begin{tabular}[c]{@{}c@{}}Scalar\\ Dot Product\end{tabular}} & Neighbor & 92.98±0.08 & 82.24±0.08 & 85.48±0.04 & 180.22 & 2.45 \\
 &  & Offset & 93.10±0.09 & 82.60±0.06 & 85.59±0.07 & 180.22 & 2.45 \\
 &  & Center, Neighbor & 92.98±0.05 & 83.03±0.08 & 85.25±0.05 & 212.99 & 4.60 \\
 &  & Center, Offset & 93.26±0.07 & 83.04±0.05 & 85.60±0.07 & 212.99 & 4.60 \\
 &  & Neighbor, Offset & 93.14±0.06 & 83.15±0.04 & 85.57±0.06 & 212.99 & 4.60 \\
 &  & Center, Neighbor, Offset & 93.20±0.08 & 83.23±0.04 & 85.74±0.06 & 245.76 & 6.75 \\ \cmidrule(l){2-8} 
 & \multirow{2}{*}{\begin{tabular}[c]{@{}c@{}}Scalar Offset\\ Dot Product\end{tabular}} & Neighbor & 93.30±0.09 & 82.77±0.06 & 85.37±0.09 & 180.22 & 2.45 \\
 &  & Center, Neighbor & 92.78±0.04 & 82.85±0.06 & 85.60±0.08 & 212.99 & 4.60 \\ \cmidrule(l){2-8} 
 & \multirow{3}{*}{\begin{tabular}[c]{@{}c@{}}Scalar\\ Addition\end{tabular}} & Neighbor & 92.65±0.07 & 82.15±0.11 & 85.37±0.06 & 180.22 & 2.45 \\
 &  & Offset & 93.38±0.05 & 82.71±0.08 & 85.54±0.04 & 180.22 & 2.45 \\
 &  & Neighbor, Offset & 93.30±0.07 & 82.94±0.09 & 85.54±0.07 & 212.99 & 4.60 \\ \cmidrule(l){2-8} 
 & \multirow{3}{*}{\begin{tabular}[c]{@{}c@{}}Scalar\\ Concat\end{tabular}} & Neighbor & 93.38±0.07 & 82.86±0.05 & 85.40±0.05 & 180.22 & 2.45 \\
 &  & Offset & 93.51±0.09 & 83.20±0.04 & 85.53±0.05 & 180.22 & 2.45 \\
 &  & Neighbor, Offset & 93.06±0.06 & 82.94±0.04 & 85.57±0.09 & 212.99 & 4.60 \\ \cmidrule(l){2-8} 
 & \multirow{2}{*}{\begin{tabular}[c]{@{}c@{}}Vector\\ Subtraction\end{tabular}} & Neighbor & \textbf{93.55±0.04} & 82.57±0.03 & 85.44±0.07 & 180.22 & 2.45 \\
 &  & Center, Neighbor & 92.94±0.04 & \textbf{83.24±0.05} & 85.50±0.07 & 212.99 & 4.60 \\ \cmidrule(l){2-8} 
 & \multirow{3}{*}{\begin{tabular}[c]{@{}c@{}}Vector\\ Addition\end{tabular}} & Neighbor & 93.14±0.09 & 82.89±0.03 & 85.74±0.07 & 180.22 & 2.45 \\
 &  & Offset & 93.06±0.07 & 82.84±0.03 & 85.43±0.11 & 180.22 & 2.45 \\
 &  & Neighbor, Offset & 93.06±0.10 & 82.95±0.05 & \textbf{85.76±0.06} & 212.99 & 4.60 \\ \bottomrule \bottomrule
\end{tabular}}
\label{table:attMethod_localFea}
\end{table}

As mentioned in \Cref{sec:localFeaAggre}, in the local-based attention module, the attention score calculation method affects the selection of features for local feature aggregation. Both aspects must be assessed simultaneously. For a more comprehensive comparison, this subsection also includes experimental results of the global-based attention module.
As shown in \Cref{table:local_aggre}, different attention score computation methods correspond to a variety of reasonable feature aggregation methods. 
Under the configuration of scale $\alpha$ = 0 and $k = 32$, the experimental results are shown in \Cref{table:attMethod_localFea}.

From the table, we can observe that for attention methods with which neighbor feature and offset feature are both applicable, when single aggregated feature is used, using offset feature achieves better performance than using neighbor feature. 
Moreover, for classification tasks, when offset feature is already used, adding neighbor feature additionally results in a modest decrease in performance. This shows that the following two principles can enable the model to perform well in classification tasks: (i) the input information contains offset features, (ii) the redundancy of the aggregated information is reduced.
However, in segmentation tasks, feature aggregation methods with higher redundancy can mostly achieve better performance. But this also brings higher FLOPs as shown in \Cref{table:attMethod_localFea}. This is a trade-off that needs to be balanced in the actual application scenarios.

In the classification task, the local-based vector offset attention $\phi^l_{v-sub}$ with neighbor feature aggregated achieves the best performance. In the segmentation task, the local-based vector offset attention $\phi^l_{v-sub}$ with center and neighbor features aggregated achieves the best performance. This provides insights for possible improvements to the existing models, e.g., both center and neighbor features can be used in HitPR \cite{hou2022hitpr} for getting better task performance.

On the other hand, we find that the global subtraction attention method also achieves decent performance with much smaller FLOPs. And the L2-norm subtraction-based attention $\phi^g_{L2}$ is overall better than the dot product self-attention.
It should be pointed out that although the FLOPs difference between the local and global attention modules is very large, with the embedding layer and task-oriented MLP head, and the actual training time difference between the two kinds of attention modes is not as large as that of FLOPs.
Under the configuration of \Cref{sec:Experiment_setting}, local-based and global-based methods take around 13 hours and 7 hours to complete the training respectively for the classification task, and 19 hours and 10 hours for the segmentation task.

\subsection{Position Encoding}
\label{sec:Position_Encoding}

\begin{table}[t]
\centering
\caption{Classification and segmentation performance obtained by selecting different position encoding methods under different attention methods. PE stands for position encoding.}
\resizebox{0.8\linewidth}{!}{
\begin{tabular}{ccccccccc}
\toprule
\toprule
\multicolumn{4}{c}{Variants} & \multirow{2}{*}{\begin{tabular}[c]{@{}c@{}}\rule{0pt}{4ex} Cls.\\ OA (\%)\end{tabular}} & \multicolumn{2}{c}{Seg.} & \multirow{2}{*}{\begin{tabular}[c]{@{}c@{}}\rule{0pt}{4ex} Params\\ (k)\end{tabular}} & \multirow{2}{*}{\begin{tabular}[c]{@{}c@{}}\rule{0pt}{4ex} FLOPs\\ (G)\end{tabular}} \\ \cmidrule(r){1-4} \cmidrule(lr){6-7}
\begin{tabular}[c]{@{}c@{}}Global/\\ Local\end{tabular} & Attention & Agg. & PE &  & \begin{tabular}[c]{@{}c@{}}Cat. mIoU\\ (\%)\end{tabular} & \begin{tabular}[c]{@{}c@{}}Ins. mIoU\\ (\%)\end{tabular} &  &  \\ \midrule
\multirow{10}{*}{Global} & \multirow{5}{*}{Dot Product} & \multirow{5}{*}{-} & - & 93.02±0.05 & 83.07±0.08 & 85.45±0.07 & 180.22 & 0.37015 \\
 &  &  & i & 93.18±0.10 & 82.90±0.08 & 85.29±0.09 & 181.38 & 0.37251 \\
 &  &  & ii & 93.06±0.07 & 83.11±0.07 & 85.35±0.07 & 180.62 & 0.37096 \\
 &  &  & iii & 93.18±0.07 & \textbf{83.22±0.04} & 85.48±0.06 & 180.99 & 0.37172 \\
 &  &  & iv & 93.22±0.03 & 83.18±0.07 & {85.53±0.07} & 181.38 & 0.37251 \\ \cmidrule(l){2-9} 
 & \multirow{5}{*}{Subtraction} & \multirow{5}{*}{-} & - & 93.43±0.07 & 83.15±0.06 & 85.47±0.06 & 180.22 & 370.15 \\
 &  &  & i & 93.38±0.08 & 83.21±0.08 & 85.55±0.07 & 181.38 & 0.37251 \\
 &  &  & ii & 93.34±0.09 & 83.12±0.09 & 85.48±0.07 & 180.62 & 0.37096 \\
 &  &  & iii & \textbf{93.51±0.04} & 83.16±0.08 & 85.60±0.06 & 180.99 & 0.37172 \\
 &  &  & iv & 93.46±0.05 & 83.19±0.05 & \textbf{85.64±0.06} & 181.38 & 0.37251 \\ \midrule
\multirow{14}{*}{Local} & \multirow{5}{*}{\begin{tabular}[c]{@{}c@{}}Scalar\\ Dot Product\end{tabular}} & \multirow{5}{*}{Offset} & - & 93.10±0.09 & 82.60±0.06 & 85.59±0.07 & 180.22 & 2.45 \\
 &  &  & i & 93.14±0.08 & 82.73±0.09 & 85.52±0.07 & 181.38 & 2.50 \\
 &  &  & ii & 93.14±0.06 & 82.85±0.07 & 85.58±0.06 & 180.75 & 2.48 \\
 &  &  & iii & 93.30±0.06 & 82.92±0.07 & 85.59±0.03 & 181.25 & 2.50 \\
 &  &  & iv & 93.22±0.06 & 83.05±0.04 & 85.77±0.03 & 181.76 & 2.50 \\ \cmidrule(l){2-9} 
 & \multirow{5}{*}{\begin{tabular}[c]{@{}c@{}}Scalar Offset\\ Dot Product\end{tabular}} & \multirow{5}{*}{Neighbor} & - & 93.30±0.09 & 82.77±0.06 & 85.37±0.09 & 180.22 & 2.45 \\
 &  &  & i & 92.25±0.07 & 82.85±0.09 & 85.43±0.10 & 181.38 & 2.50 \\
 &  &  & ii & 93.22±0.08 & 82.84±0.07 & 85.54±0.08 & 180.75 & 2.48 \\
 &  &  & iii & 93.24±0.05 & 83.15±0.05 & 85.60±0.05 & 181.25 & 2.50 \\
 &  &  & iv & 93.55±0.04 & 82.92±0.08 & \textbf{85.72±0.03} & 181.76 & 2.50 \\ \cmidrule(l){2-9} 
 & \multirow{4}{*}{\begin{tabular}[c]{@{}c@{}}Vector\\ Subtraction\end{tabular}} & \multirow{4}{*}{Neighbor} & - & 93.55±0.04 & 82.57±0.03 & 85.44±0.07 & 180.22 & 2.45 \\
 &  &  & i & 93.47±0.05 & 82.62±0.05 & 85.42±0.06 & 181.38 & 2.50 \\
 &  &  & ii & \textbf{93.56±0.05} & \textbf{83.26±0.04} & 85.55±0.05 & 181.25 & 2.50 \\
 &  &  & iii & 93.32±0.06 & 82.54±0.09 & 85.29±0.07 & 180.75 & 2.48 \\ \bottomrule \bottomrule
\end{tabular}}
\label{table:position_encoding}
\end{table}

The experimental results are reported in \Cref{table:position_encoding} and it indicates that different position encoding methods impact model performance in classifying and segmenting point cloud data. Explicit position encoding ($\delta_1$), which adds spatial coordinates directly, does improve classification accuracy to some extent, but its effectiveness is limited in complex segmentation tasks, suggesting that relying solely on absolute spatial information is insufficient for handling intricate point relationships. In contrast, implicit position encoding ($\delta_2$, $\delta_3$ and $\delta_4$) shows greater advantages, especially when combined with contextual information. By integrating rich contextual information with the Query and Key input, $\delta_3$ and $\delta_4$ significantly enhance the model's performance in both classification and segmentation tasks, emphasizing the importance of a global perspective in understanding point relationships. These findings highlight that implicit position encoding with contextual information strategy is more effective for downstream tasks.

It is worth noting that when local-based attention is used, the power of position encoding is possibly limited by the feature dimension decrease in the MLPs that are used for encoding. For example, in $\delta_2$ of \Cref{fig:local_based_position_encoding_scalar_attention)} and $\delta_3$ of \Cref{fig:local_based_position_encoding_vector_attention)}, the feature dimension of point positions has to be mapped from 3 to 1 to satisfy the architecture, resulting in less improved performance. This is also the reason why it is difficult for $\delta_2$ in the local scalar attention and $\delta_3$ in the local vector attention to improve model performance.

\section{Apply Best Practices}
\label{sec:experiments}
Through the data analysis in \Cref{sec:explore_best_practices}, The best modules under global and local attention mechanisms will be selected for testing. For module selection, we are prioritizing the best performance while also taking into account computational efficiency. Considering that downstream tasks of different complexity, we use different frameworks to handle classification and segmentation tasks.

\begin{figure}[t]
\centerline{\includegraphics[width=0.8\linewidth]{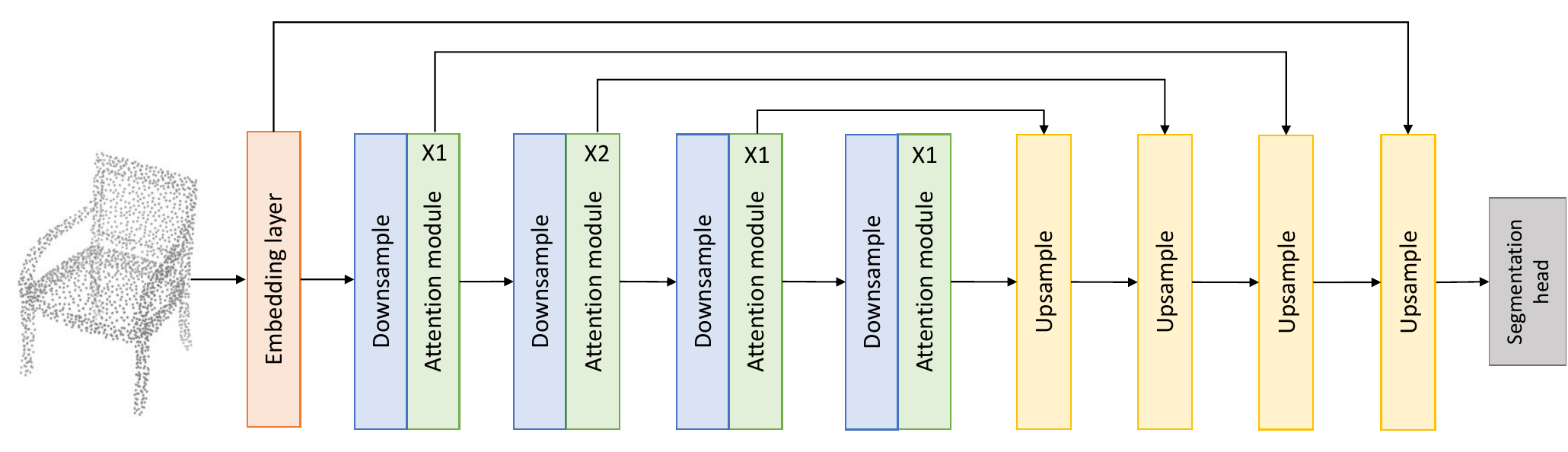}}
\caption{The final network model for the segmentation task.}
\label{fig:our_framework}
\end{figure}

\subsection{Classification}

For the classification tasks, we use the basic framework shown in \Cref{fig:basic_framework} to test on ScanObjectN and ModelNet40 benchmarks.
By applying the best practices explored before, the following two attention module variants are used: for global-based attention, we use L2-norm subtraction-based attention method $\phi^g_{L2}$ with position encoding method $\delta_3$; for local-based attention, we use subtraction-based vector attention $\phi^l_{v-sub}$ with neighbor feature aggregated and with position encoding method $\delta_2$.
The results are shown in \Cref{table:cls_ScanObjectNN_ModelNet40}. 
Our experimental results are better than or on par with the state-of-the-art methods. Moreover, please note that we achieve such superior performance with a relatively smaller number of parameters and FLOPs.

\begin{table}[t]
\centering
\caption{3D object classification performance on ScanObjectNN and ModelNet40. The parameters and FLOPs of the entire framework are also reported.}
\resizebox{0.7\linewidth}{!}{
\begin{tabular}{lccccccc}
\toprule
\multirow{2}{*}{\textbf{Method}} & \multicolumn{2}{c}{\textbf{ScanObjectNN(PB\_T50\_RS)}} & \multicolumn{2}{c}{\textbf{ModelNet40}} & \textbf{Params.} & \textbf{FLOPs} \\
 & OA(\%) & mAcc(\%) & OA(\%) & mAcc(\%) & (M) & (G) \\ \midrule
PointNet \cite{qi2017pointnet} & 68.2 & 63.4 & 89.2 & 86.2 & 3.5 & 0.9 \\
PointCNN \cite{li2018pointcnn} & 78.5 & 75.1 & 92.2 & 88.1 & 0.6 & - \\
DGCNN \cite{wang2019dynamic} & 78.1 & 73.6 & 92.9 & 90.2 & 1.8 & 4.8 \\
DeepGCN \cite{li2019deepgcns} & - & - & 93.6 & 90.9 & 2.2 & 3.9 \\
KPConv \cite{thomas2019kpconv} & - & - & 92.9 & - & 14.3 & - \\
ASSANet-L \cite{qian2021assanet} & - & - & 92.9 & - & 118.4 & - \\
SimpleView \cite{goyal2021revisiting} & 80.5±0.3 & - & 93.0±0.4 & 90.5±0.8 & 0.8 & - \\
MVTN \cite{hamdi2021mvtn} & 82.8 & - & 93.5 & 91.2 & 3.5 & 1.8 \\
PCT \cite{guo2021pct} & - & - & 93.2 & - & 2.9 & 2.3 \\
CurveNet \cite{muzahid2020curvenet} & - & - & 93.8 & - & 2.0 & - \\
PointMLP \cite{ma2022rethinking} & \textbf{85.4} & \textbf{83.9} & \textbf{94.1} & \textbf{91.3} & 13.2 & 31.3 \\ \midrule
Ours (global) & 83.1±0.4 & 80.8±0.6 & 93.8±0.1 & 90.7±0.2 & 1.93 & 3.67 \\
Ours (local) & 83.7±0.5 & 81.2±0.9 & \textbf{93.9±0.2} & \textbf{91.1±0.3} & 1.93 & 7.70
   \\ \bottomrule
\end{tabular}}
\label{table:cls_ScanObjectNN_ModelNet40}
\end{table}

\subsection{Segmentation}
A framework illustrated in \Cref{fig:our_framework} is used for the segmentation task. 
By applying the best practices explored before, the following two attention module variants are used: for global-based attention, we use L2-norm subtraction-based attention method $\phi^g_{L2}$ with position encoding method $\delta_4$; for local-based attention, we use offset dot product-based scalar attention $\phi^{l}_{dot''}$ with neighbor feature aggregated and with position encoding method $\delta_4$.
Data is progressively downsampled and processed, followed by upsampling modules to increase data resolution. The results are reported in \Cref{table:Part_segmentation_ShapeNetPart.}. 
Note that the FLOPs number is drastically decreased since the segmentation framework has multiple downsample layers.
Despite not reaching the level of state-of-the-art methods, our framework still demonstrates relatively superior performance when considering parameter scaling and computational complexity. It achieves such performance with a much smaller number of FLOPs, highlighting the efficiency and effectiveness of our attention module choices.

\begin{table}[t]
\centering
\caption{Part segmentation performance on ShapeNet Part.}
\resizebox{0.64\linewidth}{!}{
\begin{tabular}{lccccccc}
\toprule
\multirow{2}{*}{\textbf{Method}} & \multirow{2}{*}{\textbf{Cat. mIoU (\%)}} & \multicolumn{3}{c}{\multirow{2}{*}{\textbf{Ins. mIoU (\%)}}} & \textbf{Params.} & \textbf{FLOPs} \\
 &  & \multicolumn{3}{c}{} & (M) & (G)\\ \midrule
PointNet \cite{qi2017pointnet} & 80.4 & \multicolumn{3}{c}{83.7} & 3.6 & 4.9 \\
DGCNN \cite{wang2019dynamic} & 82.3 & \multicolumn{3}{c}{85.2} & 1.3 & 12.4 \\
KPConv \cite{thomas2019kpconv} & \textbf{85.1} & \multicolumn{3}{c}{86.4} & - & - \\
CurveNet \cite{muzahid2020curvenet} & - & \multicolumn{3}{c}{\textbf{86.8}} & - & - \\
ASSANet-L \cite{qian2021assanet} & - & \multicolumn{3}{c}{86.1} & - & - \\
Point Cloud Transformer \cite{guo2021pct} & 83.7 & \multicolumn{3}{c}{86.6} & 7.8 & - \\
PointMLP \cite{ma2022rethinking} & 84.6 & \multicolumn{3}{c}{86.1} & - & - \\
Stratifiedformer \cite{lai2022stratified} & \textbf{85.1} & \multicolumn{3}{c}{86.6} & - & - \\ \midrule
Ours (global) & 84.25±0.10 & \multicolumn{3}{c}{86.27±0.09} & 5.0 & 0.64 \\
Ours (local) & 84.37±0.07 & \multicolumn{3}{c}{86.36±0.05} & 5.0 & 1.39
   \\ \bottomrule
\end{tabular}}
\label{table:Part_segmentation_ShapeNetPart.}
\end{table}

\section{Conclusion}
In this paper, we conduct an extensive and fair comparative study of attention mechanisms under a unified framework and summarize some best practices in the attention module design for point cloud analysis. Furthermore, we follow these best practices and propose to use different attention modules for different downstream tasks and achieve good performance and efficiency. In summary, rethinking the attention mechanism helps to clarify the characteristic differences between different attention module variants for point cloud analysis, and provides important insights for the design and exploration of future network architectures.
\label{sec:conclusion}
\newpage


\bibliographystyle{splncs04}
\bibliography{main}

\end{document}